\documentclass[color,journal,twoside,web]{IEEEtran}

\usepackage{cite}
\usepackage{amsmath,amssymb,amsfonts}
\usepackage{algorithmic}
\usepackage{graphicx}
\usepackage{textcomp}
\usepackage{float}
\usepackage{color}
\usepackage[hidelinks]{hyperref}

\newcommand{\orcid}[1]{$^{\href{https://orcid.org/#1}{\includegraphics[width=12px]{images/orcid_icon.eps}}}$}

\usepackage{amsthm}
\theoremstyle{plain}
\newtheorem{thm}{\protect\theoremname}
\theoremstyle{definition}
\newtheorem{defn}[thm]{\protect\definitionname}

\def\BibTeX{{\rm B\kern-.05em{\sc i\kern-.025em b}\kern-.08em
    T\kern-.1667em\lower.7ex\hbox{E}\kern-.125emX}}
\providecommand{\definitionname}{Definition}
\providecommand{\theoremname}{Theorem}

\begin{document}
\title{A Spatio-temporal Attention-based Model for Infant Movement Assessment
from Videos}
\author{Binh Nguyen-Thai,~Vuong Le,~Catherine Morgan,~Nadia Badawi,~Truyen Tran and~Svetha Venkatesh
\thanks{Binh Nguyen-Thai, Vuong Le, Truyen Tran and Svetha Venkatesh are with the Applied
Artificial Intelligence Institute ($\text{A}\textsuperscript{2}\text{I}\textsuperscript{2}$),
Deakin University, Waurn Ponds, VIC 3126, Australia (E-mail: b.nguyen@deakin.edu.au; vuong.le@deakin.edu.au;
truyen.tran@deakin.edu.au; svetha.venkatesh@deakin.edu.au).}
\thanks{Catherine Morgan is with the Cerebral Palsy Alliance Research Institute, Specialty of Child \& Adolescent Health, Sydney Medical School, Faculty of Medicine \& Health, The University of Sydney, Sydney, NSW, Australia (Email: CMorgan@cerebralpalsy.org.au).}
\thanks{Nadia Badawi is with the Cerebral Palsy Alliance Research Institute, Specialty of Child \& Adolescent Health, Sydney Medical School, Faculty of Medicine \& Health, The University of Sydney, Sydney, NSW, Australia and also with the Grace Centre, The Children’s Hospital at Westmead, Sydney, NSW, Australia (Email: nadia.badawi@health.nsw.gov.au).}
\thanks{~\copyright~2021 IEEE. Personal use of this material is permitted.  Permission from IEEE must be obtained for all other uses, in any current or future media, including reprinting/republishing this material for advertising or promotional purposes, creating new collective works, for resale or redistribution to servers or lists, or reuse of any copyrighted component of this work in other works. DOI: 10.1109/JBHI.2021.3077957.}
}

\maketitle
\begin{abstract}
\noindent The absence or abnormality of fidgety movements of joints or limbs
is strongly indicative of cerebral palsy in infants. Developing computer-based
methods for assessing infant movements in videos is pivotal for improved
cerebral palsy screening. Most existing methods use appearance-based
features and are thus sensitive to strong but irrelevant signals caused
by background clutter or a moving camera. Moreover,
these features are computed over the whole frame, thus they measure
gross whole body movements rather than specific joint/limb motion.

Addressing these challenges, we develop and validate
a new method for fidgety movement assessment from consumer-grade videos
using human poses extracted from short clips. Human poses capture
only relevant motion profiles of joints and limbs and are thus free
from irrelevant appearance artifacts. The dynamics and coordination
between joints are modeled using spatio-temporal graph convolutional
networks. Frames and body parts that contain discriminative information
about fidgety movements are selected through a spatio-temporal attention
mechanism. We validate the proposed model on the cerebral palsy screening
task using a real-life consumer-grade video dataset collected at an
Australian hospital through the Cerebral Palsy Alliance, Australia.
Our experiments show that the proposed method achieves the ROC-AUC
score of 81.87\%, significantly outperforming existing competing methods
with better interpretability.

\end{abstract}

\begin{IEEEkeywords}
cerebral palsy, fidgety movement, infant movement assessment, general
movements assessment, deep learning, spatio-temporal graph convolutional
network, attention mechanism, human pose.
\end{IEEEkeywords}

\section{Introduction}

Cerebral palsy (CP) is the most common childhood physical disability
impacting movement, posture, communication, and independence in daily
activities \cite{jamapediatrics.2017.1689}. CP occurs in up to 2.5/1000
live births in high-income countries and is estimated to affect 17
million people worldwide \cite{AustralianCerebralPalsyRegister2018,Galea2019}.
The condition is caused by abnormal development or damage in the developing
brain. It is critical for an infant with CP to be diagnosed and receive
intervention as early as possible \cite{morgan2016effectiveness}.
Unfortunately, alarmingly high rates of children with CP are undiagnosed
until later childhood due to the lack of awareness or expertise. A
cheap and accurate assessment method is required.

Among early diagnostic methods of CP, the General Movements Assessment
(GMA) has been found to be cost-effective and accurate \cite{40b01a8d365e4a1386e4c22a989e3b34}.
Well-trained experts watch the videos of infants to identify the absence
or abnormality of a particular movement pattern in a developmental
state. A strong indicator for at-risk infants is the absence or sporadic
occurrence of the so-called fidgety movements \cite{einspieler2016fidgety,einspieler2015sporadic}.
Fidgety movements are small movements at moderate speed with variable
acceleration of the neck, trunks, and limbs in all directions. Such
movements typically occur in infants from approximately nine to twenty
weeks post-term. Throughout this paper, we use the term ``absent
fidgety movements'' or ``absent fidgety'' to refer to both cases:
fidgety movements are totally absent or present sporadically with
long pauses (sporadic fidgety movement) \cite{einspieler2015sporadic}.
The main drawback of GMA is that it requires skilled experts and their
time to study videos and provide the assessment. This makes the assessment
inaccessible or delayed for many infants. Therefore, there is a need
for tools that can automate the assessment of infant movements in
videos.

Modeling the movements of joints and limbs in consumer-grade videos
for fidgety movement assessment is challenging because: (i) The presence
of irrelevant signals: such signals arise from irrelevant visual artifacts,
e.g., illumination and background clutter; and camera properties.
A moving camera can produce movement signals even when the infant
does not move; a short distance from the camera to the infant will
result in greater movement signals than from longer distances. (ii)
Complex movement patterns: each joint not only has its movement patterns
but also is influenced by other joints via the body structure. (iii)
The irregularity of fidgety movements' occurrences: fidgety movements
may be stop-and-start. They occur at a joint at one time but may occur
at another joint at another time. Overall, fidgety movements only
occupy a small fraction of the whole body's movement and are weak
signals compared to other gross body movements.

Existing methods do not handle such complexity. Most current approaches
use appearance-based features \cite{Adde2010,Adde2009,Stahl2012,Rahmati2014,Stoen2017},
and thus are sensitive to irrelevant appearance artifacts and camera
properties. Therefore, the accuracies in those methods significantly
degrade when applied to consumer-grade videos. Moreover, since features
are calculated over the whole frames, they are indicative of gross
measurement of the whole body movement, not that of joints and limbs.

In this work, we develop and validate a new method that addresses
the aforementioned challenges for fidgety movement detection in consumer-grade
videos. Our method works on human poses extracted from videos. A human
pose can be represented as a time-series of 2D or 3D coordinates of
body joint locations from which relevant motion features can be computed.
Recent progress in pose estimation allows accurate extraction of 2D
poses from videos. As human poses capture only the motion profiles
of joints and limbs, our method eliminates the influence of irrelevant
signals. 

Given a video of an infant, we use a pose estimation algorithm to
extract 2D poses from its frames and generate a pose sequence (see
Section \ref{subsec:definitions}). Our method takes this pose sequence
as input and predicts the label of the infant.

\begin{figure}[tbh]
\begin{centering}
\includegraphics[scale=0.33]{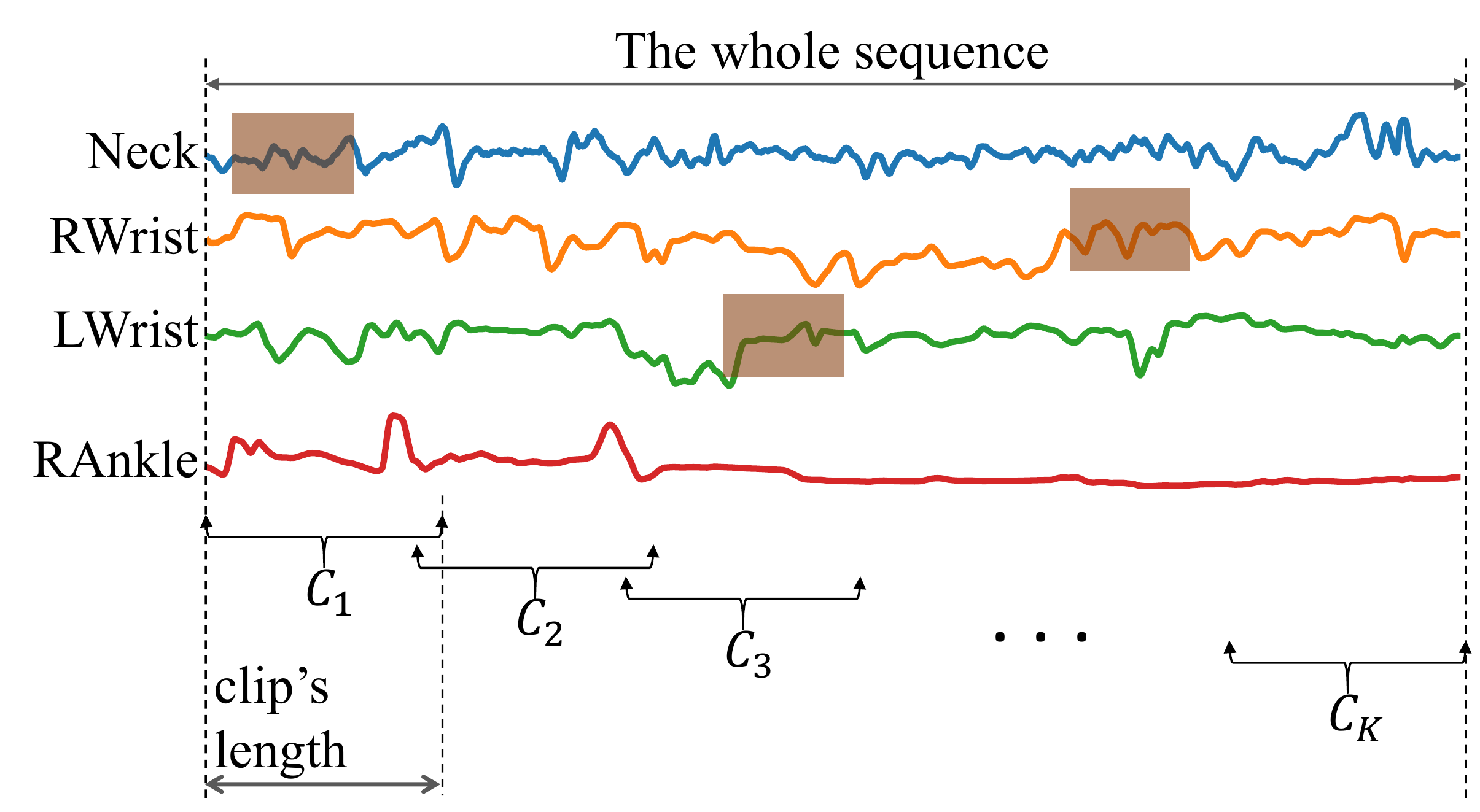}
\par\end{centering}
\caption{A pose sequence split into overlapping clips. Brown rectangles are
the places where fidgety movements happen.\label{fig:sequence_splitting}}

\end{figure}

In detail, we first split a pose sequence into a set of overlapping
fixed-length windows, referred to as clips. This is motivated by the
fact that fidgety movements occupy just a small fraction of the whole
body's movement. Splitting a long sequence into clips will allow the
model to extract useful movement patterns easier (see Fig.\ref{fig:sequence_splitting}
for an illustration). From each clip we construct a spatio-temporal
pose graph and apply a spatio-termporal graph convolutional network
(ST-GCN) \cite{stgcn2018aaai}, a model for spatio-temporal pose graph,
to learn and represent the clip. We introduce a spatio-temporal attention
mechanism on joints and clips that contain discriminative information
about fidgety movements. As a result, the proposed method can focus
on frames and body parts that contain meaningful movements, preventing
them from being buried in the whole sequence. We term our method STAM,
which stands for Spatio-temporal Attention-based Model.

We validate STAM on a dataset of 235 videos of infants who are at
around 14-15 weeks post-term age collected at an Australian hospital.
Our intensive suite of experiments shows that the proposed model significantly
outperforms existing video-based CP diagnostic models with large margins.

In summary, the main contributions of this paper are:
\begin{itemize}
\item Design and implementation of a new framework for cerebral palsy detection
from consumer-grade videos based on pose graph rather than appearance
structure; and,
\item Extensive experimentation on a real-life consumer-grade video dataset
to confirm that the proposed model achieves better accuracy than existing
baselines and is able to highlight key joints and clips.
\end{itemize}
The novelty in our framework includes: (i) a new Spatio-temporal graph
convolutional network for clips with attention to joints, and (ii)
a temporal attention mechanism for clips.

\section{Preliminaries}

\subsection{Fidgety movements\label{subsec:fidgety_movements}}

Fidgety movements (FMs) are movements of moderate speed and variable
acceleration of the neck, trunks, and limbs in all directions. FMs
are typically seen in infants from approximately 9 to 20 weeks post-term.
The absence of FMs or presence of abnormal FMs are a strong predictor
of adverse neurodevelopmental outcomes in infants. In particular,
the absence of FM between 9-20 weeks is a strong predictor of cerebral
palsy \cite{einspieler2016fidgety,einspieler2015sporadic}.

FMs often co-occur with other gross movements such as kicking, wiggling\textendash oscillating
arm movements, or antigravity movements. The temporal organization
of FMs can be divided into the following categories: (i) Continual
FMs: FMs happen frequently but are interspersed with very short pauses,
e.g., 1-2 seconds. FMs may be less obvious in the shoulders and wrists
if the infant is focused on the environment. (ii) Intermittent FMs:
The FMs are stop-and-start with longer pauses, e.g., 1-10 seconds.
(iii) Sporadic FMs: when there are more gaps than bouts of fidgety
(this is the same as absent FMs in infants between 11 and 16 weeks).
(iv) Absent FMs: No FMs can be observed, although other movements
may occur.

Thus, the occurrences of FMs are irregular in both spatial dimension
(body parts) and temporal dimension (video frames) due to the interspersion
of other movements. The detection of FMs is, therefore, a challenging
task. 

\subsection{Definitions\label{subsec:definitions}}
\begin{defn}
\textbf{Pose sequence\label{def:pose_sequence}}. The pose sequence
of $\tau$ consecutive video frames is defined as a set of $M$ multi-variate
time-series $\left\{ \mathbf{s}_{j}\right\} _{j=\overline{1,M}}$.
Here $M$ is the number of body joints, $\mathbf{s}_{j}=\left[s_{j,1},s_{j,2},\dots,s_{j,\tau}\right]$
is a sequence of motion features of the $j$-th joint with $s_{j,t}\in\mathbb{R}^{c}$
$(t=\overline{1,\tau})$, where $c$ is the dimensionality of the
motion feature vector at each joint. The pose sequence is represented
by a tensor $S\in\mathbb{R}^{M\times c\times\tau}$.
\end{defn}

\subsection{Pose graph construction \label{subsec:pose_graph_definition}}

\begin{figure}[h]
\begin{centering}
\includegraphics[scale=0.3]{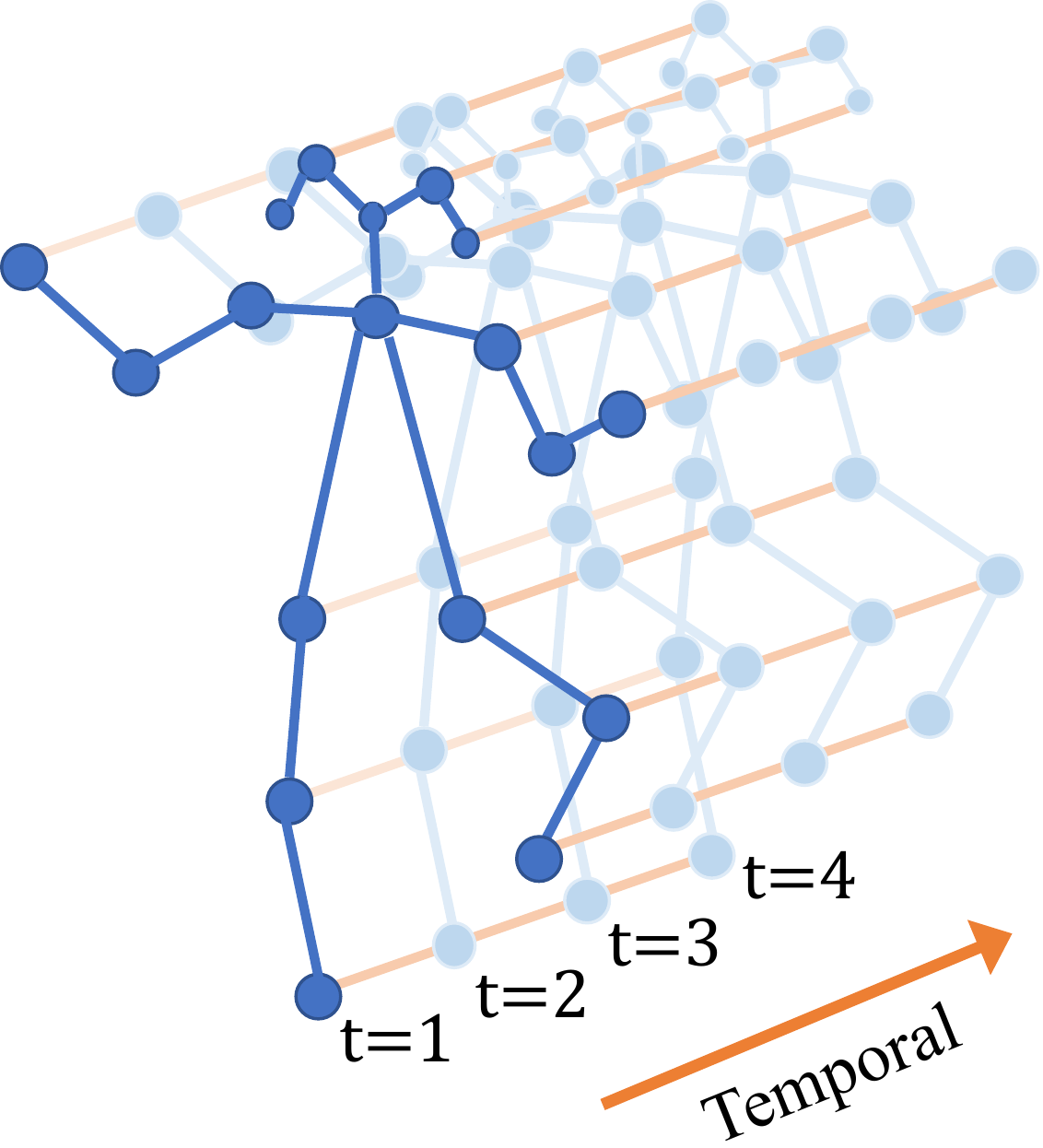}
\par\end{centering}
\caption{Example of a pose graph. Blue lines are intra-body connections. Orange
lines are inter-frame connections.\label{fig:pose_graph}}
\end{figure}

Here we present how to construct the pose graph of a pose sequence
$S\in\mathbb{R}^{M\times c\times\tau}$. The pose graph is denoted
by $G=(\mathcal{V},\mathcal{E})$, which consists of both intra-body
and inter-frame connections. The node set $\mathcal{V}=\{v_{j,t}\}$
($j=\overline{1,M}$, and $t=\overline{1,\tau}$) consists of $M\times\tau$
nodes, where $v_{j,t}$ correspond to the $j\text{-th}$ joint at
frame $t$. The feature vector of node $v_{j,t}$ is $S_{:,j,t}\in\mathbb{R}^{c}$.
The edge set $\mathcal{E}$ consists of connections constructed as
follows: intra-body connections are the limbs connecting joints according
to human body structure, while inter-frame connections are the set
of the links connecting each joint and itself on consecutive frames
(see Fig.\ref{fig:pose_graph}).

\subsection{Spatial Graph Convolutional Networks\label{subsec:gcn}}

Graph convolutional networks generalize the convolutional neural networks
(CNN) in grid data (e.g., images) to work on graph structure data
\cite{defferrard2016convolutional,estrach2014spectral,kipf2017semi}.
In the conventional CNN on 2D images, the input of the convolutional
operator is either an image or a feature map in a 2D grid, and the
output feature map is also in a 2D grid. In contrast, a convolutional
operator on graph takes node features and a graph structure as input
and generates the node-level output features. 

Convolutional graph networks can be applied to the pose graph in a
single frame to capture the coordination of the joints. Here, we follow
the Graph Convolutional Network (GCN) \cite{kipf2017semi}, a fast
and simple version of graph convolutional networks. The layer-wise
propagation rule is presented in Eq.\ref{eq:gcn_formula}.
\begin{equation}
\mathbf{Z}^{(l+1)}=\sigma\left(\tilde{\mathbf{D}}^{-\frac{1}{2}}\tilde{\mathbf{A}}\tilde{\mathbf{D}}^{-\frac{1}{2}}\mathbf{Z}^{(l)}\mathbf{W}^{(l)}\right)\label{eq:gcn_formula}
\end{equation}
Here, $\tilde{\mathbf{A}}=\mathbf{A}+\mathbf{I}_{M}$ is the adjacency
matrix of an undirected graph $G$ of $M$ nodes with added self-connections.
$\mathbf{I}_{M}$ is the $M\times M$ identity matrix, $\tilde{\mathbf{D}}_{ii}=\sum_{j}\mathbf{A}_{ij}$
is the normalized degree matrix, $\mathbf{W}^{(l)}$ a layer-specific
learnable weight matrix. $\mathbf{Z}^{(l+1)}\in\mathbb{R}^{M\times d_{l+1}}$
is the node feature map at the $l\text{-th}$ layer; $\mathbf{Z}^{(0)}\in\mathbb{R}^{M\times c}$
is the matrix of input node features. Function $\sigma\left(.\right)$
is a nonlinear activation function such as $ReLU$.

\subsection{Spatial Temporal Graph Convolutional Networks\label{subsec:st_gcn}}

Spatial Temporal Graph Convolutional Networks (ST-GCN) \cite{stgcn2018aaai}
is a graph convolutional network which takes a pose graph as input
and generates joint representations. Given a pose graph $G$, ST-GCN
generates the node feature map $\mathbf{Z}\in\mathbb{R}^{M\times d}$.
ST-GCN is used to capture both joint coordination and temporal patterns
of the movements. 

ST-GCN is a multi-layer neural network where each layer consists of
two components: (i) a GCN which is applied to every single frame to
capture the coordination of joints; and (ii) a temporal CNN, which
is applied to the output of the GCN to capture the temporal patterns.
The layer-wise propagation rule is as follows.
\begin{equation}
\mathbf{Z}^{(l+1)}=\text{Conv\ensuremath{\left(\left[\text{G}(\mathbf{Z}_{1}^{(l)}),\text{G}(\mathbf{Z}_{2}^{(l)}),\dots,\text{G}(\mathbf{Z}_{T_{l}}^{(l)})\right]\right)}}+Res
\end{equation}
Here $\mathbf{Z}^{(l+1)}\in\mathbb{R}^{M\times d_{l+1}\times T_{l+1}}$
is the output feature map of the $l\text{-th}$ layer; $\mathbf{Z}_{t}^{(l)}\in\mathbb{R}^{M\times d_{l}}$
is the $t\text{-th}$ frame of $\mathbf{Z}^{(l)}$; $\text{G}(\mathbf{Z}_{t}^{(l)})\in\mathbb{R}^{M\times d_{l+1}}$
is the feature map obtained by applying GCN to frame $t$ of $\mathbf{Z}^{(l)}$
(see Eq.\ref{eq:gcn_formula}); $\mathbf{Z}^{'}\in\mathbb{R}^{M\times d_{l+1}\times T_{l}}$
is a tensor obtained by stacking $\text{G}(\mathbf{Z}_{t}^{(l)})$
($t=\overline{1,T_{l}}$). Conv is the temporal CNN component which
consists of a 2D CNN with kernel size=$(t\text{\_}kernel,1)$ and
stride=$(t\text{\_}stride,1)$. $Res$ is a residual layer which is
a 2D CNN with kernel size=$(1,1)$ and stride=$(t\text{\_}stride,1)$.
Here, $t\text{\_}kernel$ and $t\text{\_}stride$ are hyper-parameters
specifying the temporal kernel size and temporal stride, respectively

The output of the last layer, $\mathbf{Z}^{(L)}\in\mathbb{R}^{M\times d\times T_{L}}$,
is fed to an average pooling over the third dimension ($T_{L}$) to
obtain the joint feature map $\mathbf{Z}\in\mathbb{R}^{M\times d}$.

\section{Proposed Method}

\subsection{Pipeline overview}

\begin{figure}[H]
\begin{centering}
\includegraphics[scale=0.36]{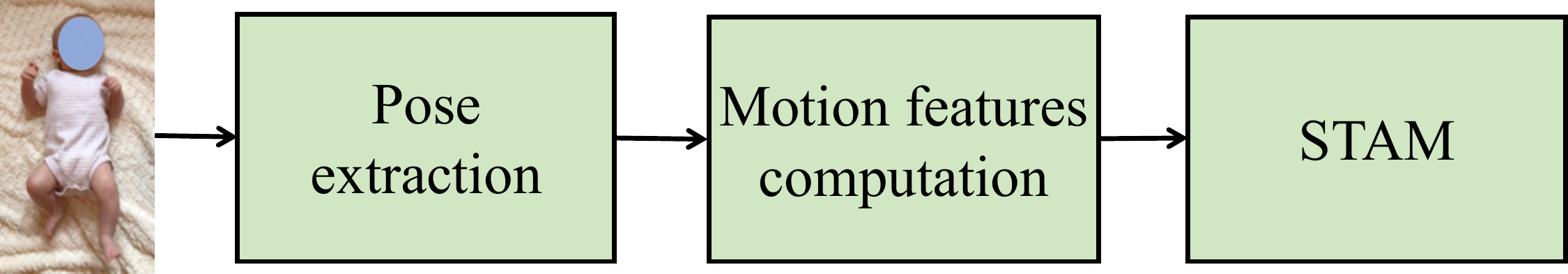}
\end{centering}
\caption{Pipeline of the proposed method.\label{fig:pipeline_overview}}
\end{figure}

The pipeline of our method is presented in Fig.\ref{fig:pipeline_overview}.
First, we use a pose estimation algorithm to extract poses from video
frames in the form of a set of time-series of 2D joint coordinates.
Then we compute other motion features such as the velocities and accelerations
of joints to form a pose sequence. These pose sequences are the input
to STAM for training a classifier to predict the label of a pose sequence.
We will detail the components.

\subsection{Pose extraction}

\begin{figure}[H]
\begin{centering}
\includegraphics[scale=0.33]{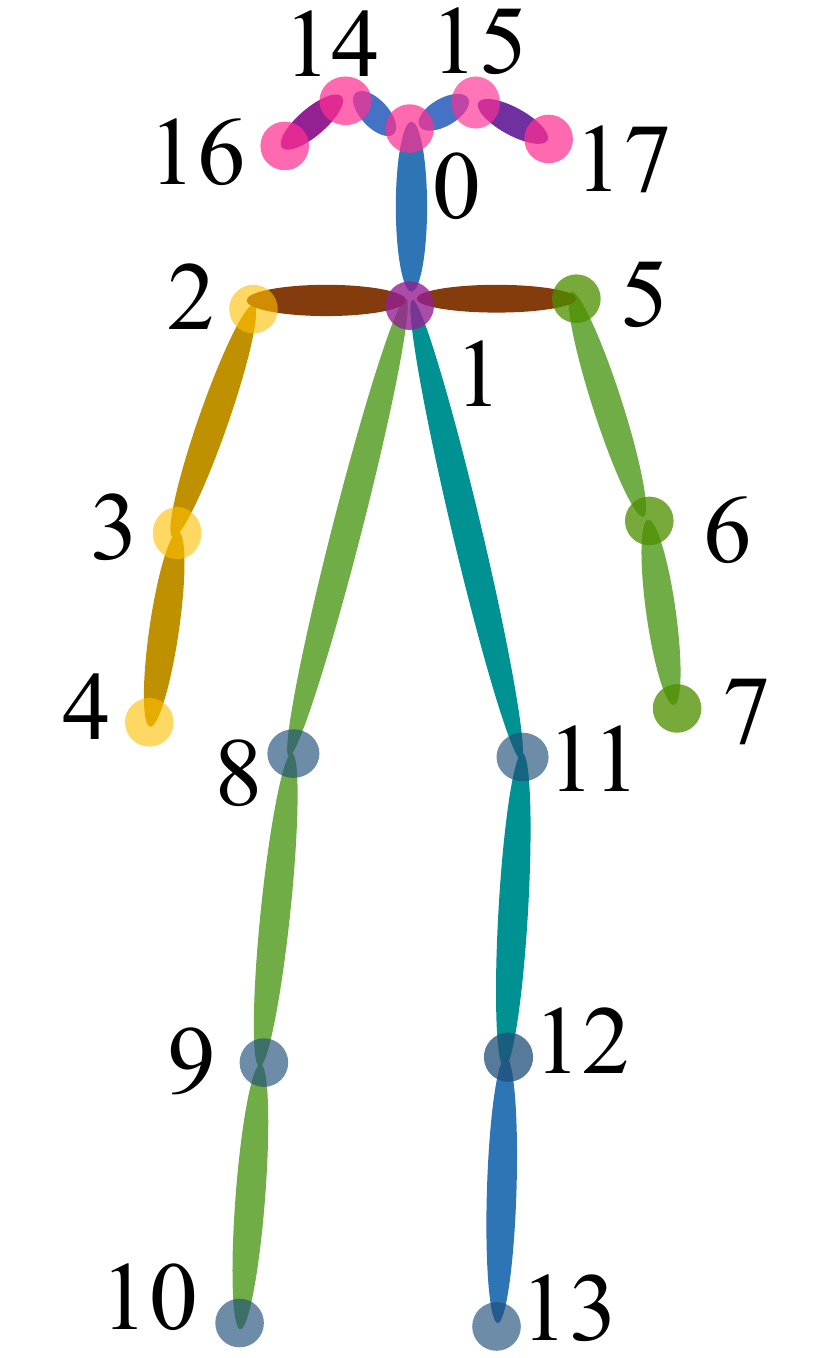}
\par\end{centering}
\caption{Pose layout of a single frame. Joint indices: 0: nose, 1: neck, 2:
right shoulder, 3: right elbow, 4: right wrist, 5: left shoulder,
6: left elbow, 7: left wrist, 8: right hip, 9: right knee, 10: right
ankle, 11: left hip, 12: left knee, 13: left ankle, 14: right eye,
15: left eye, 16: right ear, 17: left ear.\label{fig:skeleton_layout}}
\end{figure}

\textbf{Pose estimation}: We
used OpenPose \cite{cao2017realtime}, a pose estimation algorithm,
to extract poses from video clips. OpenPose is a deep learning-based
model that is trained on annotated image data to identify 2D positions
of joints and limbs of humans in images. There are multiple options
for the number of joints and the layouts of the pose to be extracted.
In this work, we choose to use the layout with 18 joints, as presented
in Fig.\ref{fig:skeleton_layout}. We include eyes and ears as they
are relevant to the movements of the neck. By applying OpenPose, we
estimated the positions of 18 joints for every frame.

OpenPose is originally trained using adult data,
thus is suboptimal for infants due to the difference in body size
and appearance. Therefore, retraining OpenPose using an annotated
infant data is needed. This task was done in \cite{Chambers2019},
with a pre-trained model provided. We use this pre-trained model for
the pose estimation in our data.

\begin{figure}[tbh]
\begin{centering}
\includegraphics[scale=0.32]{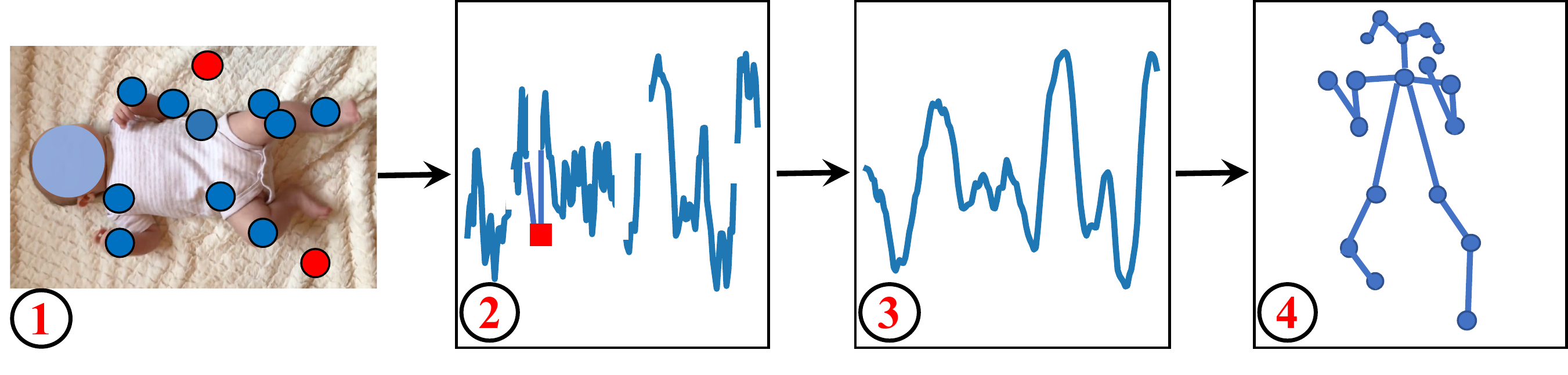}
\par\end{centering}
\caption{Pipeline of the data pre-processing. (\textbf{1}) the input video
with estimated joint locations (the dots). The data contains missing
data (the right ankle, right wrist, left shoulder), outliers (red
dots). (\textbf{2}) the time-series of joint coordinates with missing
data (white space) and outlier (the red dot). (\textbf{3}) the time-series
obtained after interpolating, removing outlier, and smoothing. (\textbf{4})
the pose obtained after standardizing.\label{fig:data_preprocessing}}
\end{figure}

\textbf{Preprocessing}.:The estimated
pose contained missing data (i.e., some joints are not detected) and
outliers (e.g., background clutter that is inaccurately detected as
joints). We perform a pre-processing process which includes: (i) \emph{data
imputation}: we imputed missing data using a linear
interpolation to the raw time-series for each joint; (ii) \emph{outlier
removal}: we removed the outliers by using a rolling-median
filter with a smoothing window of 0.5 seconds; and (iii) \emph{signal
smoothing}: we performed data smoothing by using
a rolling-mean filter with a smoothing window of 0.5 seconds (see
Fig.\ref{fig:data_preprocessing}).

\begin{figure*}[tbh]
\begin{centering}
\includegraphics[scale=0.4]{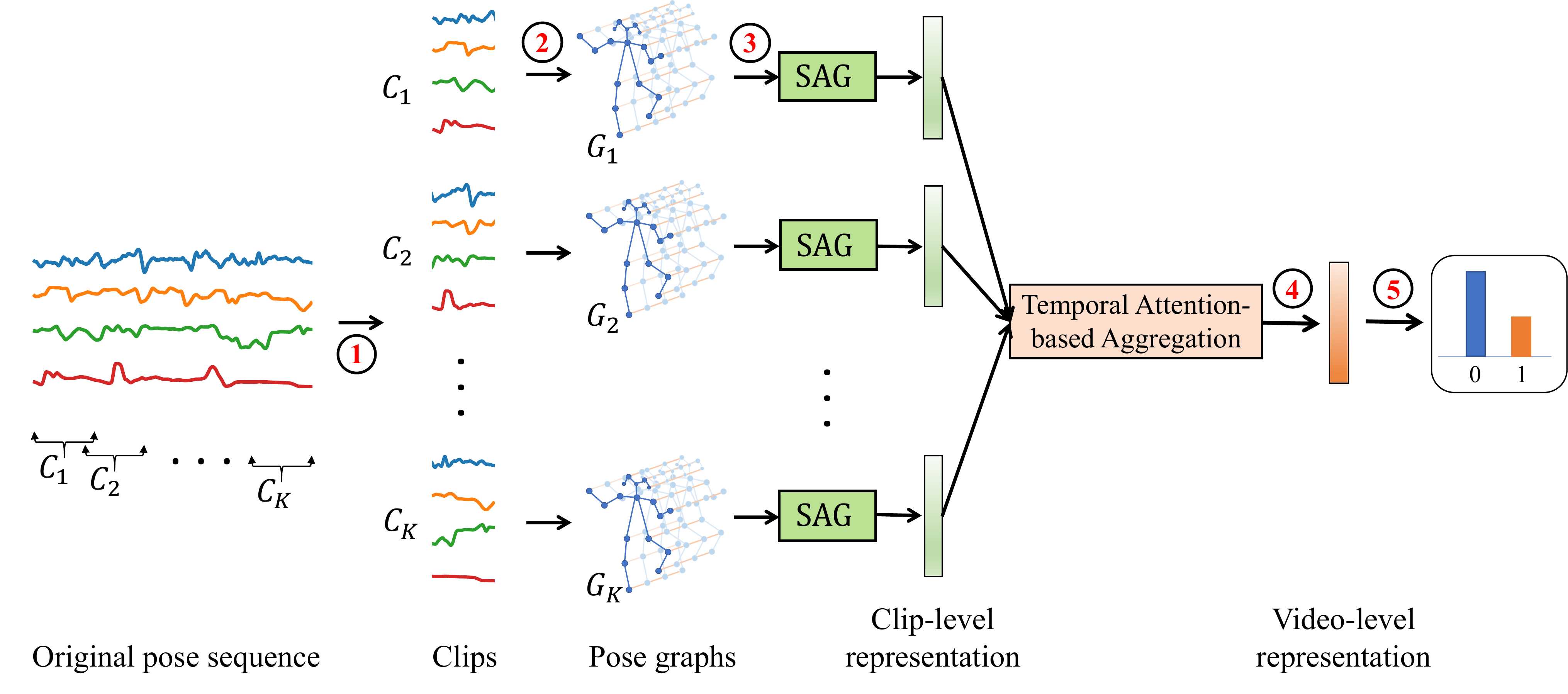}
\par\end{centering}
\caption{High-level architecture of STAM. \textbf{Step 1}: split the pose sequence
into overlapping clips. \textbf{Step 2}: construct the pose graph
of each clip. \textbf{Step 3}: generate clip-level representations
using Spatial Attention Graph Convolutional Network (SAG) (defined
in Section \ref{subsec:stam}). \textbf{Step} 4: compute video-level
representation using temporal attention. \textbf{Step }5: make predictions.\label{fig:stam_architecture}}
\end{figure*}

In consumer-grade datasets, the videos are taken
in various conditions such as camera viewpoints, distances from the
camera to the infant. To remove the effect of these factors, we follow
the process outlined in \cite{Chambers2019} to convert the poses
to body-centered coordinates and normalize them to the same view and
scale. The process includes: (i) rotating the upper body's joints
around the center point of the shoulders; (ii) rotating the lower
body's joints around the center point of the hips; and (iii) normalizing
the body size using the trunk length (i.e., the distance between the
neck and the center of the hips) as a reference distance: set the
trunk length to 1 and translate the joint coordinates to get the ratios
between the trunk length and the distances between joints to the neck
remained (see Fig.\ref{fig:data_preprocessing}).

\subsection{Motion features computation \label{subsec:motion_feature_computation}}

From the time-series of joint coordinates, we further
compute other high-order motion features at each time step, which
are important for modeling the motions. Those motion features are
the velocities, the accelerations, and the travel distances of joints
at each frame. Velocities and accelerations are smoothed using a rolling-mean
filter with a smoothing window of 0.25 seconds. The motion features
of all frame will be stacked to form a pose sequence $S\in\mathbb{R}^{M\times7\times T}$
of the video in which $S_{j,:,t}=\left[x_{j,t},y_{j,t},v_{j,t}^{(x)},v_{j,t}^{(y)},a_{j,t}^{(x)},a_{j,t}^{(y)},d_{j,t}\right]^{\top}$
is the motion feature vector of the $j\text{-th}$ joint at frame
$t$. Here $(x_{j,t},y_{j,t})$ is the 2D coordinate; $\left(v_{j,t}^{(x)},v_{j,t}^{(y)}\right)$
and $\left(a_{j,t}^{(x)},a_{j,t}^{(y)}\right)$ are the velocities
and accelerations over $x$ and $y$ axis; and $d_{j,t}$ is the travel
distance of the $j\text{-th}$ joint at frame $t$.

\subsection{STAM: A \underline{S}patio-\underline{t}emporal \underline{A}ttention-based \underline{M}odel}
\label{subsec:stam}

Given the pose sequences extracted from the videos,
we now proceed to construct our model to learn and represent the dynamics
and coordinations of the joints and limbs for making predictions.
We are given a set of sequence-label pairs $\mathcal{D=}\left\{ \left(S_{i},y_{i}\right)\right\} _{i=\overline{1,N}}$
of $N$ videos, where $S_{i}\in\mathbb{R}^{M\times c\times T}$ and
$y_{i}\in\{0,1\}$ are the pose sequence and the label of the $i\text{-th}$
video, respectively. Here $y_{i}=1$ if normal fidgety movements are
present, and $y_{i}=0$ otherwise (e.g., cerebral palsy). We are interested
in learning a classifier $f:\mathbb{R}^{M\times c\times T}\rightarrow\{0,1\}$
that predicts the label of an unseen pose sequence.

Our method, STAM, is motivated by the irregularity in the occurrences
of fidgety movements (section \ref{subsec:fidgety_movements}). Fidgety
movements may occur at some joints in some consecutive video frames,
however, the joints and frames are unknown. So we can break down a
long sequence into overlapping clips (windows) where fidgety movements
only occur at some joints in some of these clips. STAM will learn
to select the joints and clips that contain discriminative information
about fidgety movements.The high-level architecture
of STAM is shown in Fig.\ref{fig:stam_architecture}. Given the pose
sequence $S_{i}$, we split it into $K$ fixed-length clips $\left(C_{i,1},C_{i,2},\dots,C_{i,K}\right)$
where $C_{i,k}$ $(k=\overline{1,K})$ is a pose sequence of length
$w$. From each $C_{i,k}$, we construct a pose graph using the procedure
described in Section \ref{subsec:pose_graph_definition}, resulting
in a set of $K$ pose graphs $\left(G_{i,1},G_{i,2},\dots,G_{i,K}\right)$.
For simplicity, we omit the subscription $i$ from the notations when
no confusion arises, presenting a general infant. First, the pose
graphs are fed to a Spatial Attention Graph Convolution Network (SAG),
a component of STAM, which learns joint representations and aggregates
them to generate clip-level representations, via a \emph{spatial
attention} mechanism. Then, clip-level representations
are aggregated by a \emph{temporal attention} mechanism to form the 
video-level representation, which will be used
for the prediction.

\begin{figure}[t]
\begin{centering}
\includegraphics[scale=0.4]{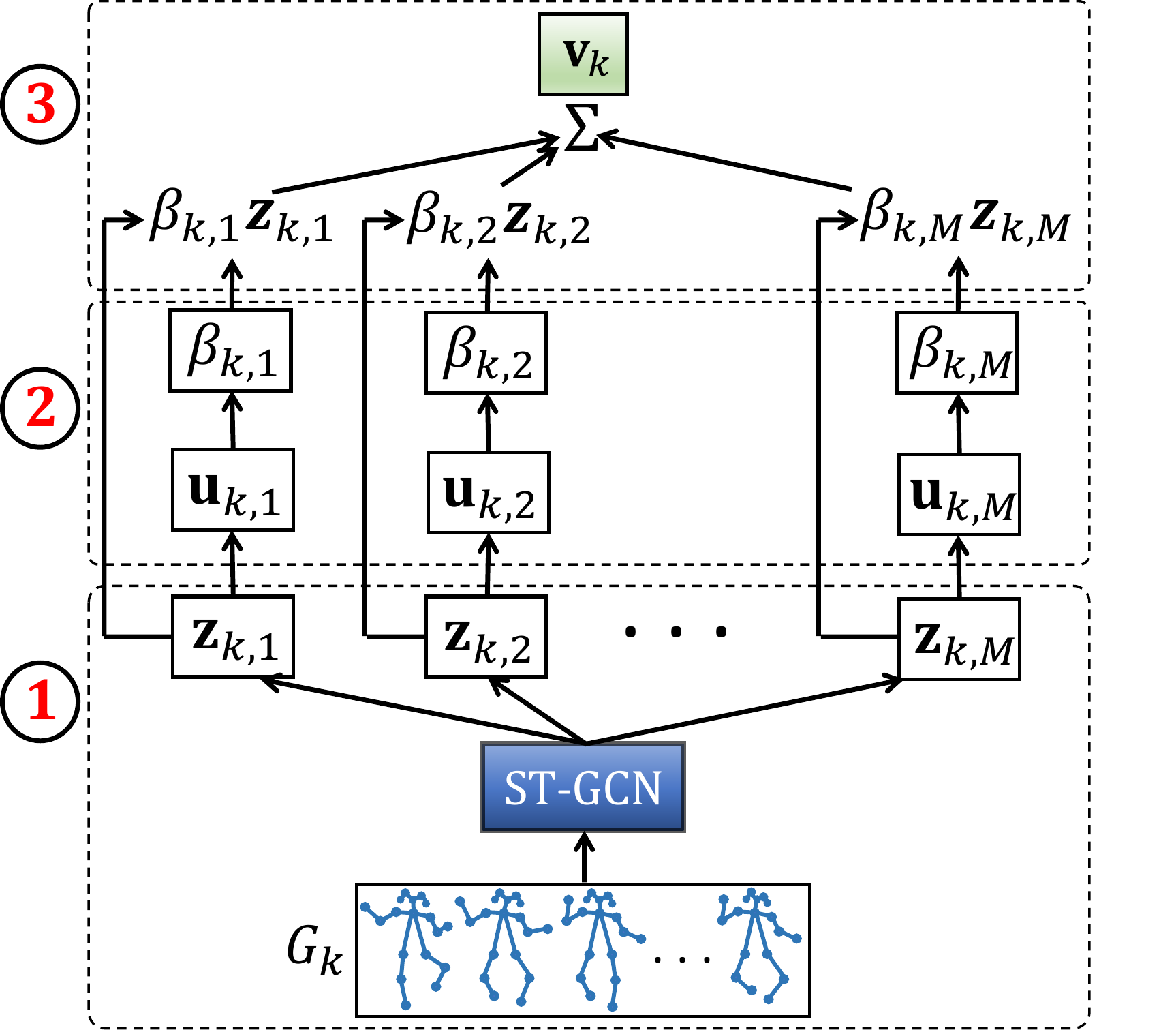}
\end{centering}
\caption{Architecture of the Spatial Attention Graph Convolutional Network
(SAG). \textbf{Step 1}: generate joint representations using ST-GCN,
\textbf{Step 2}: generate spatial attention weights, \textbf{Step
3}: aggregate joints to compute the clip-level representation.\label{fig:sag}}
\end{figure}

\textbf{\underline{S}}\textbf{patial}\textbf{\underline{A}}\textbf{ttention
}\textbf{\underline{G}}\textbf{raph Convolutional Network (SAG)}. We now detail SAG,
a network that learns the representation of a pose graph with attention
to joints. SAG is applied to clips for learning to generate clip-level
representations. SAG extends ST-GCN (see Section \ref{subsec:st_gcn})
with a spatial attention layer stacked on the top of the ST-GCN (see
Fig.\ref{fig:sag}). This layer generates the attention weights for
joints and aggregates joint representations to obtain the clip-level
representation.

In detail, a pose graph $G_{k}$ $(k=\overline{1,K})$
is first fed into a ST-GCN to generate joint representations $\mathbf{z}_{k,j}\in\mathbb{R}^{d}$
$(j\in\overline{1,M})$.
\begin{equation}
\mathbf{z}_{k,1},\mathbf{z}_{k,2},...,\mathbf{z}_{k,M}=\text{ST-GCN}\left(G_{k}\right)
\end{equation}

Then, joint representations will be aggregated to
obtain the clip-level representation. We propose to use the weighted
average of the joint representations as the aggregation function,
where the weights are generated by a neural network. Given a set of
joint representations $\left\{ \mathbf{z}_{k,1},\mathbf{z}_{k,2},\dots,\mathbf{z}_{k,M}\right\} $
of the $k\text{-th}$ clip, we propose the following spatial attention-based
aggregation function.
\begin{equation}
\mathbf{v}_{k}=\sum_{j=1}^{M}\beta_{k,j}\mathbf{z}_{k,j}
\end{equation}
where the spatial attention weight $\beta_{k,j}$ is calculated as
follows:
\begin{equation}
\mathbf{u}_{k,j}=tanh(\mathbf{W}_{u}\mathbf{z}_{k,j})
\end{equation}
\begin{equation}
\beta_{k,j}=\frac{\text{exp}\left(\mathbf{w}_{\beta}^{\top}\mathbf{u}_{k,j}\right)}{\sum_{j^{'}}\text{exp}\left(\mathbf{w}_{\beta}^{\top}\mathbf{u}_{k,j^{'}}\right)}
\end{equation}
where $\mathbf{W}_{u}\in\mathbb{R}^{d_{u}\times d}$ and $\mathbf{w}_{\beta}\in\mathbb{R}^{d_{u}}$
are learnable parameters.

\textbf{Temporal attention-based aggregation.}

Clip-level representations will be aggregated to
obtain the video-level representation. We propose to use the weighted
average of the clip-level representations as the aggregation function,
where the weights are generated by a neural network. Given a set of
clip-level representations $\left\{ \mathbf{v}_{1},\mathbf{v}_{2},\dots,\mathbf{v}_{K}\right\} $
of a pose sequence $S$, we propose the following temporal attention-based
aggregation function.
\begin{equation}
\mathbf{c}=\sum_{k=1}^{K}\alpha_{k}\mathbf{v}_{k}
\end{equation}
 where the temporal attention weight $\alpha_{k}$ is calculated as
follows:
\begin{equation}
\mathbf{h}_{k}=tanh(\mathbf{W}_{h}\mathbf{v}_{k})
\end{equation}
\begin{equation}
\alpha_{k}=\frac{\text{exp}\left(\mathbf{w}_{\alpha}^{\top}\mathbf{h}_{k}\right)}{\sum_{k^{'}}\text{exp}\left(\mathbf{w}_{\alpha}^{\top}\mathbf{h}_{k^{'}}\right)}
\end{equation}
where $\mathbf{W}_{h}\in\mathbb{R}^{d_{h}\times d}$ and $\mathbf{w}_{\alpha}\in\mathbb{R}^{d_{h}}$
are learnable parameters.

\textbf{Prediction}: We use the
video-level representation $\mathbf{c}$ of the pose sequence $S$
to predict the true class $y=\{0,1\}$ as follows:
\begin{equation}
\hat{y}=P(y=1|S)=\text{\ensuremath{\sigma}}(\mathbf{w}^{\top}\mathbf{c}+b)\label{eq:prediction_formula}
\end{equation}
where $\sigma(.)$ is the sigmoid function, $\mathbf{w}\in\mathbb{R}^{d}$
and $b\in\mathbb{R}$ are learnable parameters.

\textbf{Loss function}: We use
the cross-entropy loss as the objective function for learning the
model parameters.
\begin{equation}
\mathcal{L}\left(\mathcal{D}\right)=-\frac{1}{N}\sum_{i=1}^{N}\left(y_{i}\log\hat{y}_{i}+(1-y_{i})\log(1-\hat{y}_{i})\right)\label{eq:loss_function}
\end{equation}

\subsection{Interpreting STAM}

Here, we present a method to interpret the behavior of STAM by analyzing
the attention weights. The availability of attention weights at joint-
and clip-level allows us to gain the insight about the contributions
of joints at different segments of the video rather than only at the
whole-sequence level.

We rewrite the Eq.(\ref{eq:prediction_formula}) as follows.

\begin{eqnarray}
P(y=1|S) & = & \sigma\left(\mathbf{w}^{\top}\sum_{k=1}^{K}\alpha_{k}\mathbf{v}_{k}+b\right)\nonumber \\
 & = & \sigma\left(\mathbf{w}^{\top}\sum_{k=1}^{K}\alpha_{k}\sum_{j=1}^{M}\beta_{k,j}\mathbf{z}_{k,j}+b\right)\label{eq:interpretation}\\
 & = & \sigma\left(\sum_{k=1}^{K}\sum_{j=1}^{M}\alpha_{k}\beta_{k,j}\mathbf{w}^{\top}\mathbf{z}_{k,j}+b\right)\nonumber 
\end{eqnarray}

Eq.(\ref{eq:interpretation}) allows us to compute the contribution
of each joint at each clip to the video's label. Specifically, $\alpha_{k}\beta_{k,j}\mathbf{w}^{\top}\mathbf{z}_{k,j}$
can be interpreted as the contribution of the $j$-th joint at the
$k$-th clip. Since $\mathbf{w}$ is shared across the infants, we
can omit $\mathbf{w}$ and use $\alpha_{k}\beta_{k,j}$ as the importance
of the joints. We assume that high value of $\alpha_{k}\beta_{k,j}$
should be assigned to the places (joints and clips) where fidgety
movements occur.

\section{Validation}

\subsection{Data}

We used videos of 235 infants who are at around 14-15 weeks post-term
age, provided by one hospital in Australia. Videos were taken as part
of another study and were used with permission. These videos were
primarily captured by parents using the BabyMoves App \cite{kwong2019baby}
via a smartphone in the infant\textquoteright s natural environment.
Instructions for taking the video include using a plain background,
having the infant lightly clothed and not sucking on a pacifier, or
interacting with people or toys. Amongst the videos used for this
analysis, there was considerable variability in lighting, background,
the distance and orientation to the camera (camera view).The videos
vary in resolution: 360x480 (147 videos), 480x360 (9 videos), 480x640
(3 videos), 480x720 (76 videos). The length of each video is approximately
4500 frames, equivalent to 2.5 minutes (30 frames per second). Of
the videos used in this dataset, 200 were classified as normal fidgety,
and the remaining 35 as absent fidgety.

\subsection{Pose extraction}

We extract the human pose using a pre-trained model for infant pose
estimation \cite{Chambers2019}. The model was trained using a dataset
of infant images with ground-truth labels of joint positions. Since
this pre-trained model was carefully tuned on infant videos, we did
not perform further evaluation on our dataset. Here we summarize the
performance of the pose estimation based on the results reported in
\cite{Chambers2019}. The performance of the pose estimation was evaluated
using three metrics: (i) Root Mean Square Error (RMSE): the RMSE between
the ground-truth joints and the estimated joints; (ii) Precision:
the proportion of estimated joints that were present in the ground-truth
joints; and (iii) Recall: the proportion of ground-truth points that
are detected by the pose estimation algorithm. To account for differences
in scale, the RMSE is normalized by the size of the ground-truth bounding-box
around the infant. As reported in \cite{Chambers2019}, the RMSE in
bounding-box unit is 0.02, the precision is 0.92, and the recall is
0.94. With this accuracy, the pre-trained model is applicable to extract
joint positions in our infant videos.

\subsection{Experimental settings}

\textbf{Train/test split}. We split data into training and validation
sets with ratio 8:2, using stratified random training/validation split.
Stratified random split is used to retain the ratio of positive and
negative examples in training and validation sets. To avoid the effect
of data randomness, we perform re-shuffling and splitting 10 times,
and report the average accuracy on the validation sets in all experiments.
The number of positive (i.e., absent fidgety) and negative (i.e.,
normal fidgety) examples are: 28 positive and 160 negative examples
for the training set; and 7 positive and 40 negative examples for
the validation set.

\textbf{Voting based-prediction}. To increase the amount of data,
we split each pose sequence into 4 sub-sequences using a sliding window
with length 1000 frames and 200 overlapping frames, resulting in 112
positive and 640 negative examples for the training set and 28 positive
and 160 negative examples for the validation set. The original sequence
is considered positive if there exists one of its sub-sequences predicted
as positive.

\textbf{Metrics}. We use Area Under the ROC Curve (ROC-AUC) of comparing
$\hat{y}_{i}$ with the true label $y_{i}$ as the metric to evaluate
the discrimination power of the proposed model. ROC-AUC is the most
popular method to measure model's discrimination.

\subsection{Baseline methods}

We compare STAM with the following baselines:
\begin{itemize}
\item \textbf{MotionImage} \cite{Stoen2017}: This method uses appearance-based
features. First, the videos are cropped to the smallest bounding boxes
containing the infants. Then the motion images are calculated by subtracting
consecutive frames. The quantity and the centroid of the motions are
calculated from the motion images. Q-mean (mean of the quantity) and
C-std (standard deviation of the centroid) of the motions are used
as the scores for the classification.
\item \textbf{Logistic Regression}: This is a logistic regression model
which uses the feature vectors computed following \cite{Chambers2019}
as input features. These features include the averages, medians, entropies,
and deviations of the coordinates, velocities, accelerations. We normalize
these feature vectors to zero-mean and unit-variance. The resulting
vectors are used to train the logistic regression.
\item \textbf{Conv2D-HOJD2D} \cite{McCay2020}: This method applies a 2D
convolutional neural network (Conv2D) to the Histogram of Joint Displacement
2D (HOJD2D) computed from human poses. The movement of each joint
is represented by tracking its displacement at every frame. The displacements
are organized into 16 bin histogram to form a feature vector in $\mathbb{R}^{2\times16}$.
The feature vectors of 12 limb joints are stacked to form a HOJD2D
matrix of size $\mathbb{R}^{24\times16}$ which is used as the input
of the Conv2D.
\item \textbf{LSTM (Long-short term memory)}: To demonstrate the modeling
power of STAM, we implement a baseline that uses the same pose sequence,
without graph structure, as input. For this purpose, we use LSTM,
which is standard for modeling sequential data. The input of this
LSTM is a multivariate time-series of size $\mathbb{R}^{7M\times T}$,
where each variable is a motion feature as described in Section \ref{subsec:motion_feature_computation}.
Logistic regression is applied to the top hidden layer for predicting
the label.
\item \textbf{ST-GCN }(Spatial Temporal Graph Convolutional Network) \cite{stgcn2018aaai}:
This is a graph convolutional network for spatio-temporal data such
as human pose.
\end{itemize}

\subsection{Implementation details}

For STAM, we use overlapping clips of length of 30 frames with 10
frames overlapping. We use a ST-GCN with three convolutional layers
whose output channels are 64, 128, 256, respectively. Three convolutional
layers use temporal kernel size $t\text{\_}kernel=9,9,9$ and temporal
stride $t\text{\_}stride=1,2,2$, respectively. We put a batch norm
layer and a non-linear layer using $ReLU$ function before the 2D
CNN of the temporal CNN. We put a batch norm layer and a dropout layer
with dropout rate=0.3 after the 2D CNN of the temporal CNN. The dimensionality
of the clip-level representations and video-level presentation is
$d=256$. For Conv2D-HODJ2D, we use the architecture described in
\cite{McCay2020} with two convolutional layers. Each convolutional
layer consists of a 2D convolutional layer with $kernel\text{\_}size=3$
and $stride=1$ followed by a max-pooling layer with $kernel\text{\_}size=3$
and a dropout layer with $dropout\text{\_}rate=0.3$. For the LSTM,
we use a network that consists of three hidden layers with dimensionality
of 128. We use Adam \cite{kingma2014adam} as the optimizer with learning
rate $lr=0.0001$ and weight decay $0.0001$ for all methods. We terminate
the training after 800 epochs and report the results at the epoch
giving the best validation accuracy. Source code is available at https://github.com/nguyenthaibinh/stam.

\subsection{Result comparison}

We report the ROC-AUC of all methods on Table \ref{tab:roc_auc}.
As clearly seen, STAM significantly outperforms all competing methods.
We also have the following observations.

\begin{table}[tbh]
\begin{centering}
\caption{ROC-AUC of STAM and competing methods. $(^{*})$: methods that use
appearance-based motion features.\label{tab:roc_auc}}
\par\end{centering}
\centering{}%
\begin{tabular}{|l|c|}
\hline 
\textbf{Methods} & \textbf{ROC-AUC}\tabularnewline
\hline 
\hline 
MotionImage-Q-mean \cite{Stoen2017}\textsuperscript{{*}} & 0.5352 (0.0869)\tabularnewline
\hline 
MotionImage-C-std \cite{Stoen2017}\textsuperscript{{*}} & 0.5465 (0.0835)\tabularnewline
\hline 
Logistic Regression & 0.6581 (0.1142)\tabularnewline
\hline 
Conv2D-HOJD2D \cite{McCay2020} & 0.6771 (0.0639)\tabularnewline
\hline 
LSTM & 0.6925 (0.0341)\tabularnewline
\hline 
ST-GCN & 0.7675 (0.0632)\tabularnewline
\hline 
STAM (proposed) & \textbf{0.8187 }(0.0377)\tabularnewline
\hline 
\end{tabular}
\end{table}

(1) The pose features-based methods significantly outperform appearance
features-based methods. This is expected because appearance-based
features include irrelevant information as our dataset contains a
lot of noise from background clutter and camera shakiness. This suggests
that using pose features is clearly better for consumer-grade videos.
(2) Among pose features-based methods, Conv2D-HOJD2D, LSTM, and STAM
significantly outperform Logistic Regression. This demonstrates the
effectiveness of deep learning-based techniques in modeling the motions
over just a simple model on hand-crafted features. (3) We observe
the advantage of modeling the motion dynamics as LSTM, ST-GCN, and
STAM outperform Conv2D-HOJD2D, which only captures the gross movements
over the whole sequence. (4) The superiority of STAM and ST-GCN over
LSTM confirms the advantage of taking into account the coordination
of joints via graph structure. (5) The most encouraging result is
that the proposed model STAM significantly outperforms ST-GCN. This
confirms the effectiveness of the spatio-temporal attention mechanism
which learns to focus on body parts and video frames that are meaningful
for discriminating fidgety and non-fidgety movements.

\subsection{ROC curve for practical use}

\begin{figure}[tbh]
\begin{centering}
\includegraphics[scale=0.37]{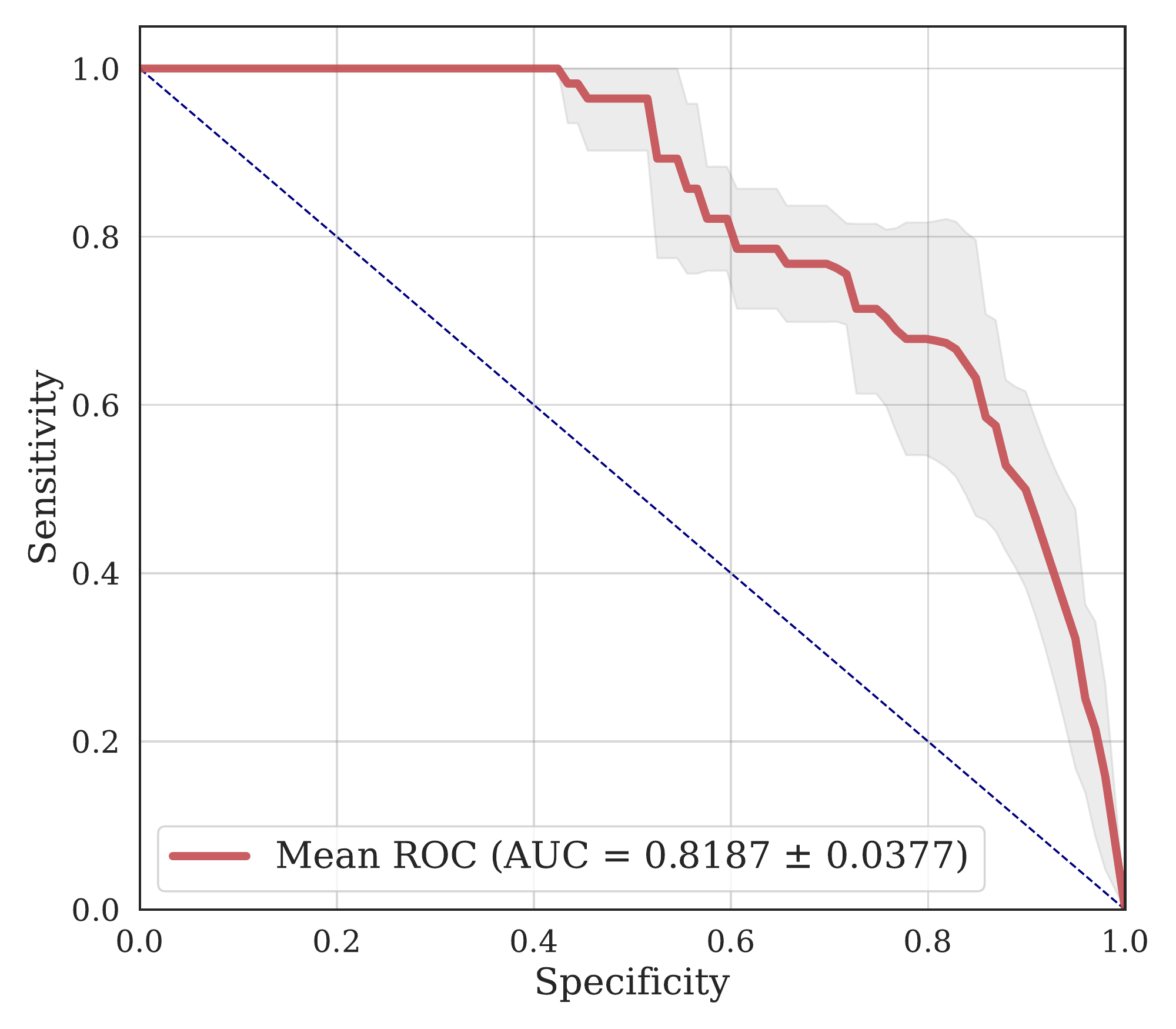}
\par\end{centering}
\caption{ROC curve of the proposed model.\label{fig:mean_roc}}
\end{figure}

We analyze the extent to which the proposed model can help practitioners
in screening infants at risk. Fig. \ref{fig:mean_roc} shows the ROC
curve, which presents the relationship between sensitivity and specificity.
Based on this, end users can adjust the classification threshold to
have the optimal sensitivity and specificity. For example, a threshold
can be chosen to obtain a sensitivity of 80\%, at a specificity of
62\%.

\subsection{Qualitative analysis}

\begin{figure}[h]
\begin{centering}
\includegraphics[scale=0.285]{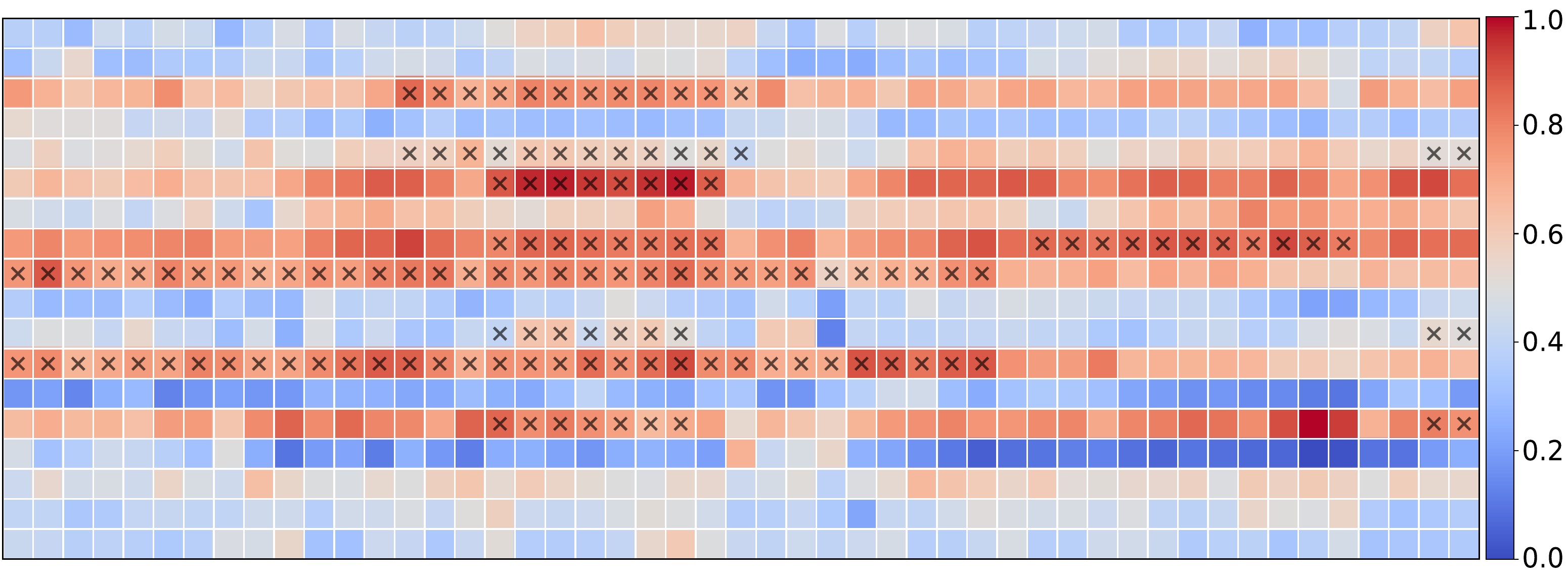}
\par\end{centering}
\caption{Attention weights of a healthy infant. Each row corresponds to a joint
where the row index follows the joint index presented in Fig.\ref{fig:skeleton_layout}.
Each column corresponds to a clip where the column index is the clip\textquoteright s
index.\label{fig:joint_attention_weight_negative}}
\end{figure}

To demonstrate the interpretability of the proposed attention mechanism,
we present the attention weights of a healthy infant in Fig. \ref{fig:joint_attention_weight_negative}.
Here each row corresponds to a joint and each column corresponds to
a clip. The value of cell $(j,k)$ is $\alpha_{k}\beta_{k,j}$ after
being normalized to $\left[0,1\right]$, and is expressed by the cell's
color. The darker the red color, the higher the cell's value; the
darker the blue color, the lower the cell's value, as shown in the
color bar on the right of the grid. A high value of a cell indicates
a high contribution of the corresponding joint and clip to the video's
classification. To verify the behavior of the attention weights, we
asked an expert to annotate the places where fidgety movements happen.
The cells corresponding to these places are marked with ``$\times$''.
We can see that there is a substantial matching between the with highest
values and ``$\times$'' marker. Basically, high attention weights
are assigned to the cells where fidgety movements occur (marked with
``$\times$''), while low attention weights are assigned to the
cells where fidgety movements do not happen. 

That observation confirms that the proposed model not only attains
high predictive accuracy but also is able to highlight body parts
and frames that contain discriminative information about fidgety movements.

\begin{figure}[h]
\begin{centering}
\includegraphics[scale=0.31]{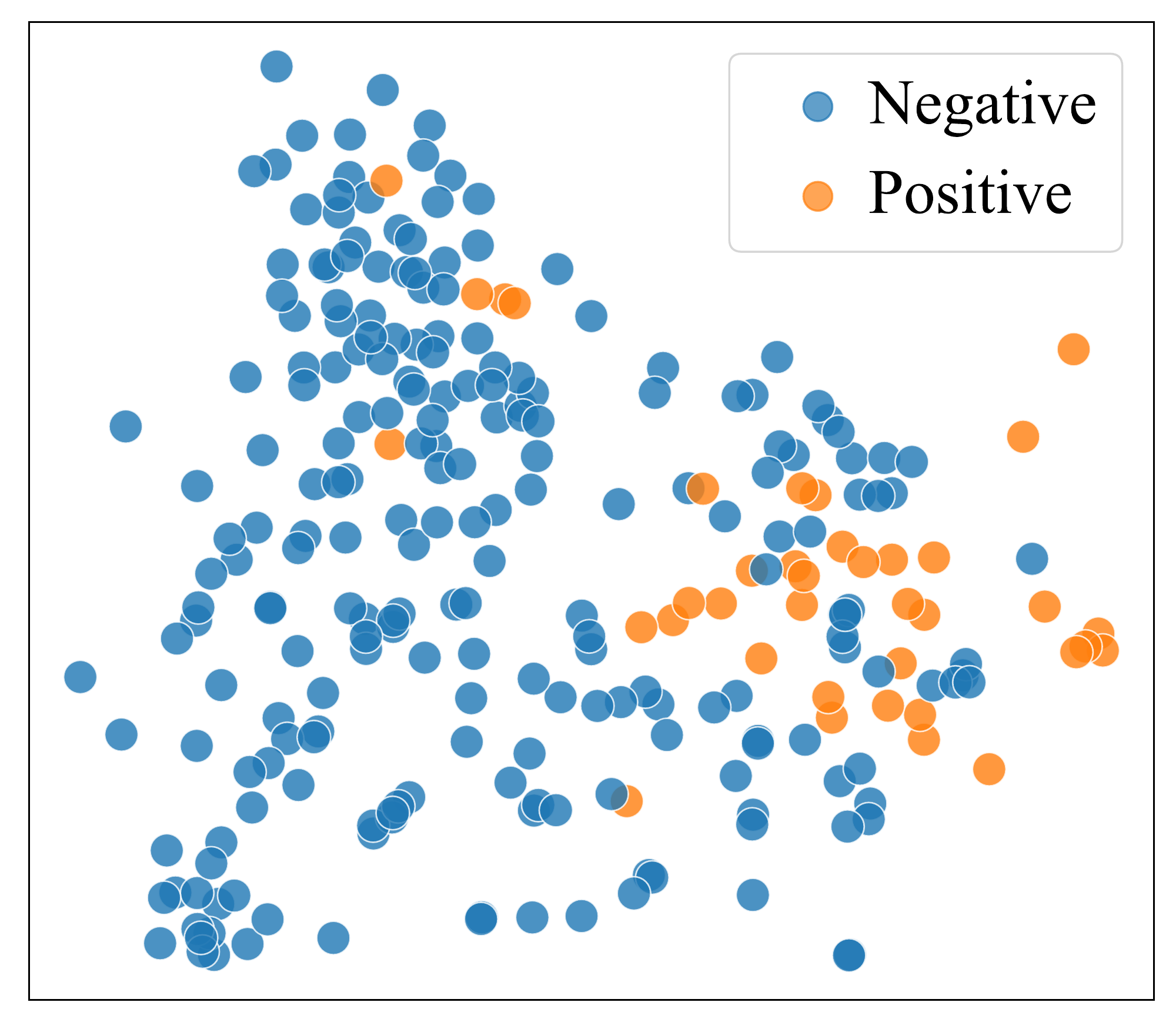}
\par\end{centering}
\centering{}\caption{Plot of the feature space using t-SNE.\label{fig:tsne_feature_space}}
\end{figure}

Fig.\ref{fig:tsne_feature_space} plots the video-level representations
$\mathbf{c}_{i}$ $(i=\overline{1,N})$ using t-Distributed Stochastic
Neighbor Embedding (t-SNE) \cite{van2008visualizing}. We can see
that most of the positive infants (orange dots) are well separated
from normal infants (blue dots), showing the effectiveness of the
proposed model in learning good discriminative representations.

\subsection{Ablation study}

To gain more insight of the proposed model, we performed an ablation
study by stripping STAM components. The results are reported in Table
\ref{tab:ablation_study}. Overall, STAM-full (the the proposed model)
outperforms its ablated variants, confirming the effectiveness of
the proposed model. Details are as follows.

\begin{table}[H]
\begin{centering}
\begin{tabular}{|l|c|}
\hline 
\textbf{Model} & \textbf{ROC-AUC}\tabularnewline
\hline 
STAM-full & \textbf{0.8187 }(0.0377)\tabularnewline
\hline 
\textbf{Attention} & \tabularnewline
\textcolor{white}{\textemdash{} }STAM-w/o attention & 0.7704 (0.0605)\tabularnewline
\textcolor{white}{\textemdash{} }STAM-w/ spatial attention & 0.7853 (0.0641)\tabularnewline
\textcolor{white}{\textemdash{} }STAM-w/ temporal attention & 0.7951 (0.0528)\tabularnewline
\hline 
\textbf{Clip modeling} & \tabularnewline
\textcolor{white}{\textemdash{} }LSTM-clip & 0.7363 (0.0742)\tabularnewline
\hline 
\end{tabular}
\par\end{centering}
\caption{The accuracies of STAM and its simplified variants.\label{tab:ablation_study}}
\end{table}

\textbf{Impact of the attention mechanism}. We study the following
STAM variants: (i) STAM-w/o attention: a variant that does not use
the attention mechanism. Clip-level representations are the means
of the joint representations. The video-level representation is the
mean of its clip representations; (ii) STAM-w/ spatial attention:
a variant that uses only spatial attention. Clip-level representations
are the weighted averages of the joint representations using spatial
attention weights $\beta_{j}$. The video-level representation is
the mean of its clip representations; and (iii) STAM-w/ temporal attention:
a variant that uses only temporal attention. Clip-level representations
are the means of the joint representations. The video-level representation
is the weighted average of its clip representations using temporal
attention weights $\alpha_{k}$.

We have the following observations: (1) Attention-based variants outperform
non-attention variant. This result indicates that both spatial and
temporal attention can add value in enhancing the predictive accuracy.
(2) The improvement is amplified in STAM-full, when both spatial attention
and temporal attention are used. This result confirms the superiority
of STAM by selecting meaningful body parts and video frames simultaneously
through the spatio-temporal attention.

\textbf{Impact of pose graph modeling}. We study LSTM-clip, a variant
of STAM that replaces the SAG with a multi-variate LSTM. Each clip
is modeled by an LSTM to generate the clip-level representation. The
LSTM's input is a multi-variate time-series which is exactly the same
as the input of SAG, except the body graph structure.

The superiority of STAM confirms that modeling the pose graphs will
handle the complexity of the movements better than considering the
movements of joints independently.

\subsection{Time for the video assessment}

After training, STAM can be used to assess an unseen infant's video.
The time needed to process and predict the video's label is the sum
of the time for: (i) human pose extraction from the video; (ii) preprocessing
and motion feature computation; and (iii) label prediction. For a
video of length 1000 frames (equivalent to 33 seconds), the total
time needed is approximately 50 seconds (see Table \ref{tab:assessement_time}
for the details).

\begin{table}[tbh]
\begin{centering}
\caption{Details of the time needed to assess a video.\label{tab:assessement_time}}
\par\end{centering}
\centering{}%
\begin{tabular}{|l|c|}
\hline 
\textbf{Step} & \textbf{Time (sec)}\tabularnewline
\hline 
\hline 
Human pose extraction & 34\tabularnewline
\hline 
Motion feature computation & 15\tabularnewline
\hline 
Label prediction & 1\tabularnewline
\hline 
\textbf{Total} & 50\tabularnewline
\hline 
\end{tabular}
\end{table}

\section{Discussion}

We have presented STAM, a method to identify infants at risk of cerebral
palsy via video-based infant movement assessment. We extract human
poses from the videos and model the movements of infants using a spatio-temporal
graph neural network. We use a spatio-temporal attention mechanism
to select body parts and video frames which contain discriminative
features. STAM offers the following advantages over existing methods:
(i) it operates on human poses extracted from videos, thus it is robust
to irrelevant visual signals such as background clutter; (ii) it learns
to extract meaningful motion features via spatio-temporal graph neural
networks and the attention mechanism; and (iii) it provides a better
interpretability through the spatial and temporal attention weights.

\subsection{Significance}

This is the first time spatio-temporal graph convolutional networks
and attention mechanisms are used with human pose sequences for cerebral
palsy prediction. Since the human pose captures only the movements
of joints and limbs, it is robust against irrelevant visual artifacts
in videos. By normalizing the pose of every frame into the same view
and scale, the method is independent of camera properties such as
camera shakiness or camera angle. This allows the method to work well
on consumer-grade videos. To verify the robustness of STAM, we exclude
the videos captured in bad conditions from the test set and see whether
the accuracy increases. The excluded videos are in one of the following
cases: (i) the videos which contain background clutter, (ii) the videos
which were captured by moving cameras, and (iii) the videos which
were captured from side angles. We did not observe any significant
change in the accuracy for all cases. In other words, including videos
captured in bad conditions does not affect the model's accuracy.

In terms of modeling, by applying the spatio-temporal graph convolutional
neural networks for modeling the movements, the proposed method can
capture the complex motion dynamics of joints and limbs. The attention
mechanism enables the model to amplify weak-but-important signals,
prevent them from being buried in strong but irrelevant signals.

With those capabilities, the proposed method is expected to aid the
cerebral palsy screening in many ways. First, it enables objective
movement assessment. Judgment by humans may be subjective, so clinicians
can cross-check with the result of our method to obtain objective
decisions. Second, it can help reduce the specialist assessment burden
and increase the number of infants that can be assessed as the method
can process the videos taken at home by parents without the need of
waiting to take the videos in the clinical environment. Finally, as
our method uses human pose, it is a privacy-preserving technique,
thus it enables data sharing between cohorts and hospitals for training
the models without the risk that the infants are identified.

\subsection{Comparison to other works on infant movement assessment}

\textbf{Wearable sensor-based methods}. Wearable sensors and 3D motion
capture have been used to measure infant movements \cite{Fan2012,Heinze2010,Kanemaru2014,Karch2012,Karch2008,Philippi2014,Rahmati2016,Redd2019,Singh2010}.
Sensor-based measurements are objective and quantitative, and algorithms
can generate risk assessments. However, these measurements have been
restricted to laboratory and clinical settings, thus hindering the
ability to evaluate infants in their natural environments. Moreover,
sensor-based methods can be costly and time-intensive to develop and
implement. 

\textbf{Video-based methods}. Another line of methods for infant movement
assessment is video-based. These methods use consecutive frame differencing
to estimate movement features which are either the amount and centroid
of motion \cite{Adde2010,Adde2009,Stahl2012} or optical flow \cite{Rahmati2014,Stoen2017}.
These features are used directly as scores or the input of a classifier.
However, these pixel-based features are sensitive to irrelevant visual
artifacts such as background clutter and illumination. Moreover, the
features are extracted from the entire frame, over the whole sequence
without focusing on meaningful body parts and frames.

\textbf{Human pose-based methods}. Recently, there have been several
works that use human poses extracted from video clips. In \cite{Chambers2019},
the authors extract infant poses from video clips then compute hand-crafted
features such as velocities, accelerations, and entropies of the joints.
They then compute the surprise score as the abnormal score for the
movement. In \cite{McCay2020}, the authors proposed to use Histograms
of Joint Orientation and Histograms of Joint Displacement. Those histogram
features are used as the input of a convolutional neural network to
train a movement classifier. However, since the features are calculated
over the whole sequence, important frames and body parts are easy
to be buried in the other part of the video. In addition, these methods
cannot model the joint coordinations and dynamics of the movements.

\subsection{Limitations and future work}

There still remains some limitations of the proposed method. First,
STAM depends on the accuracy of the pose estimation algorithm, which
may introduce noise or is sometimes inaccurate due to the occlusion
of limbs. This can be addressed by fusing visual features guided by
the human pose, e.g., the optical flow around joints and limbs. Second,
since STAM is a supervised learning model, it needs annotated data
which is usually hard to acquire. This can be solved by using self-supervised
learning techniques to learn good representations from large datasets
of unlabeled videos, e.g., learning to reconstruct masked joints.

\section{Conclusion}

We have developed and validated STAM, a new method for video-based
infant movement assessment for cerebral palsy screening. The method
uses human poses extracted from videos and is based on deep learning
techniques using graph structures and spatio-temporal attention mechanisms.
Experimental results confirm the competitive advantages of STAM on
cerebral palsy screening. The key findings are: (i) Pose sequences
are strong signals that retain motion information of joints and limbs
whilst ignoring irrelevant, distracting visual artifacts; (ii) Representing
pose sequences as spatial-temporal graphs and modeling the movements
by spatial-temporal graph convolutional networks work well as they
capture the coordination and temporal patterns of joints and limbs;
(iii) Attention mechanisms are useful for identifying discriminative
patterns which are distributed over time and body parts.

\emph{Ethical Approval}: The study was approved by the Mater Misericordiae
Ltd Human Research Ethics Committee (MML HREC) (EC00332) in Australia
(Review number: HREC/14/MHS/188).

\bibliographystyle{IEEEtran}
\bibliography{references/wearable_sensor_based,references/rgb_video_based,references/pose_feature_based,references/background}

\end{document}